\pdfoutput=1

\documentclass[11pt]{article}

\usepackage[final]{acl}

\usepackage{times}
\usepackage{latexsym}

\usepackage[T1]{fontenc}

\usepackage[utf8]{inputenc}

\usepackage{microtype}

\usepackage{inconsolata}

\usepackage{graphicx}
\usepackage{url}

\usepackage{makecell}
\usepackage{colortbl}
\usepackage{xcolor}
\usepackage{soul}
\usepackage{multicol}
\usepackage{multirow}
\usepackage{blindtext}
\usepackage{dirtree}
\usepackage{tikz}

\usepackage{pbox}
\usepackage{chngpage}
\usepackage{minitoc}
\usepackage{longtable}

\usepackage{amsmath, amsfonts, amsthm, stackrel, amssymb} 

\usepackage{tcolorbox}
\usepackage{caption}
\usepackage{subcaption}
\usepackage{hyperref}

\usepackage{graphicx}
\usepackage{multicol}
\usepackage{multirow}
\usepackage{blindtext}
\usepackage{dirtree}
\usepackage{tikz}
\usepackage{booktabs} 

\usepackage{newfloat}
\usepackage{listings}
\usepackage{enumitem}

\usepackage{cleveref}
\usepackage[linesnumbered,ruled,vlined]{algorithm2e}
\usepackage[thicklines]{cancel}
\usepackage{tablefootnote}

\newcommand{\ourmethod}{{FRAG}}
\newcommand{\llamaChat}[1]{{Llama2-#1B}}
\newcommand{\llamaInstruct}[1]{{Llama3-#1B}}
\newcommand{\llamaChatFull}[1]{{Llama-2-#1b-chat-hf}}
\newcommand{\llamaInstructFull}[1]{{Llama-3-#1B-Instruct}}
\newcommand{\KGQA}{{KGQA}}

\title{FRAG: A Flexible Modular Framework for Retrieval-Augmented Generation based on Knowledge Graphs}

\author{
	Zengyi Gao$^{1,2}$\thanks{Equal contribution.}, Yukun Cao$^{1,2}$\footnotemark[1],
    Hairu Wang$^{1,2}$, Ao Ke$^{1,2}$, Yuan Feng$^{1,2}$, Xike Xie$^{1,2}$\thanks{Corresponding author.}, S Kevin Zhou$^{1,3}$ \\
	$^1$University of Science and Technology of China, China\\ $^2$Data Darkness Lab, MIRACLE Center, USTC, China\\ $^3$MIRACLE Center, USTC, China\\
	\footnotesize\texttt{\{gzy02,ykcho,hrwang00,sa21225249,yfung\}@mail.ustc.edu.cn}, 
	\footnotesize\texttt{\{xkxie,skevinzhou\}@ustc.edu.cn}
}

\begin{document}

\maketitle

\begin{abstract}

To mitigate the hallucination and knowledge deficiency in large language models (LLMs), Knowledge Graph (KG)-based Retrieval-Augmented Generation (RAG) has shown promising potential by utilizing KGs as external resource to enhance LLMs reasoning.
However, existing KG-RAG approaches struggle with a trade-off between flexibility and retrieval quality.
Modular methods prioritize flexibility by avoiding the use of KG-fine-tuned models during retrieval, leading to fixed retrieval strategies and suboptimal retrieval quality.
Conversely, coupled methods embed KG information within models to improve retrieval quality, but at the expense of flexibility.
In this paper, we propose a novel flexible modular KG-RAG framework, termed FRAG, which synergizes the advantages of both approaches.
FRAG estimates the hop range of reasoning paths based solely on the query and classify it as either simple or complex.
To match the complexity of the query, tailored pipelines are applied to ensure efficient and accurate reasoning path retrieval, thus fostering the final reasoning process.
By using the query text instead of the KG to infer the structural information of reasoning paths and employing adaptable retrieval strategies, FRAG improves retrieval quality while maintaining flexibility.
Moreover, FRAG does not require extra LLMs fine-tuning or calls, significantly boosting efficiency and conserving resources.
Extensive experiments show that FRAG achieves state-of-the-art performance with high efficiency and low resource consumption.%


\end{abstract}

\section{Introduction}
%
%
%
%
%

Large language models (LLMs) excel in various NLP tasks but are prone to hallucinations \shortcite{hong2023faithful} and errors \shortcite{wang2023knowledge} when answering questions that require knowledge beyond  training data. These limitations undermine the trustworthiness of LLMs and raise security concerns \shortcite{rathee2024}.
To mitigate these issues, retrieval-augmented generation (RAG) \shortcite{guu2020,ColBERT,min2023} has been developed, which dynamically retrieves information from external sources during the {\it retrieval phase}, while the {\it generation phase} leverages this retrieved data to improve generation quality.
Previous RAG methods \shortcite{HyDE,Rewriting,gao2024,StepBackRAG} relying on unstructured data often struggle with capturing relevant knowledge and may introduce noise, hindering effective reasoning.
In response to these challenges, knowledge graphs (KGs) are increasingly being integrated into RAG (i.e., {\it KG-RAG}) as external knowledge sources \shortcite{UniKGQA, G-Retriever, GRAG,jiang2024kg, GNN-RAG}. KGs offer editable and explicit knowledge  in a structured format, clarifying the context and multi-level interrelations among entities. KG-RAG retrieves ``reasoning paths'' that are relevant to the query, providing concise and structured contextual information to enhance the reasoning  ability of LLMs \shortcite{talmor2018,talmor2019,petroni2021,yu2023,pan2023unifying}.

\begin{figure}[t]
    \centering
    \includegraphics[width=0.99\columnwidth]{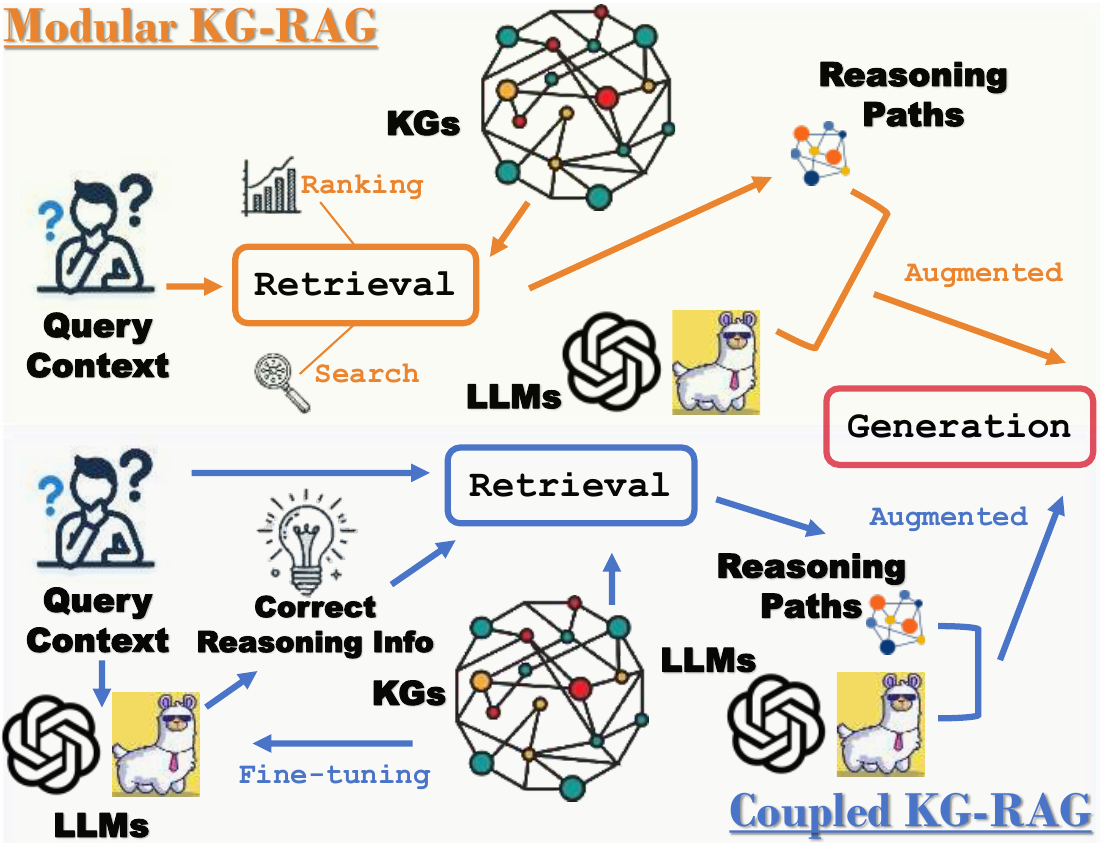}
    \caption{Modular and Coupled KG-RAG Frameworks}
    \label{fig:kgrag}
\end{figure}

There are two lines of research on KG-RAG frameworks, as shown in Figure~\ref{fig:kgrag}, depending on their ways of integration with LLMs during the retrieval phase, resulting in either {\it modular} or {\it coupled} KG-RAG.
The former separates the retrieval process from LLMs, prioritizing isolation, flexibility, and scalability, while the latter tightly integrates KGs, improving generation quality but at the expense of increased complexity and reduced flexibility and scalability.

Modular KG-RAG frameworks \shortcite{KB-BINDER,KD-CoT,knowledgeaugmented,ToG}, aligning with conventional RAG principles, are developed to keep independence between the retrieval phase and KG-fine-tuned LLMs or specific models, while meeting essential criteria of isolation, flexibility, and scalability. This isolation prevents interference with LLMs’ internal reasoning. Flexibility is achieved through seamless plug-and-play integration with external KG sources, and scalability allows these frameworks to effectively handle large KGs and complex queries. This is accomplished by using traditional algorithms for reasoning path search \shortcite{G-Retriever,ToG,ToG2.0} and ranking \shortcite{PathRerank,ToG2.0}, without requiring any additional fine-tuning to incorporate KGs information prior to reasoning.
Despite the advantages, modular KG-RAG frameworks face challenges, due to their lack of prior knowledge about KGs and correct reasoning paths. This limitation impedes effective adjustment of searching and ranking parameters, leading to inferior generation quality \shortcite{knowledgeaugmented,KB-BINDER,KD-CoT}.
For example, research like ToG \shortcite{ToG} highlights that fixed search parameters can result in redundancy in simple tasks with short reasoning paths or the omission of critical details in more complex tasks with longer paths, ultimately weakening the reasoning capabilities of LLMs.

In contrast, coupled KG-RAG frameworks, while offering enhanced retrieval of reasoning paths, come with significant overhead due to the extensive fine-tuning of LLMs with embedded KG information \shortcite{RoG,GNN-RAG,ChatKBQA}.
For instance, RoG \shortcite{RoG} not only use KGs as an external knowledge base but also fine-tune LLMs or train the specific models with KGs information, enabling the generation of ``relation paths'' grounded in KGs as retrieval templates tailored to the query context.
While this method enhances retrieval and generation quality, it compromises key advantages offered by modular KG-RAG frameworks, such as the isolation of the retrieval process, flexibility in integrating various external sources, and scalability to KGs of different sizes \shortcite{knowledgeaugmented,KB-BINDER}, not to mention the substantial expense involved in fine-tuning LLMs.

This leads us to our research goal: {\it Can we develop a solution that combines the strengths of both frameworks}? Specifically, can we implement lightweight pre-computation to obtain sufficient information to guide the settings of ranking and searching parameters during the retrieval phase, thereby enhancing generation quality while maintaining isolation, flexibility, and scalability—without the extensive overhead associated with fine-tuning LLMs for specific KGs?

\begin{figure}[t]
    \centering
    \includegraphics[width=0.99\columnwidth]{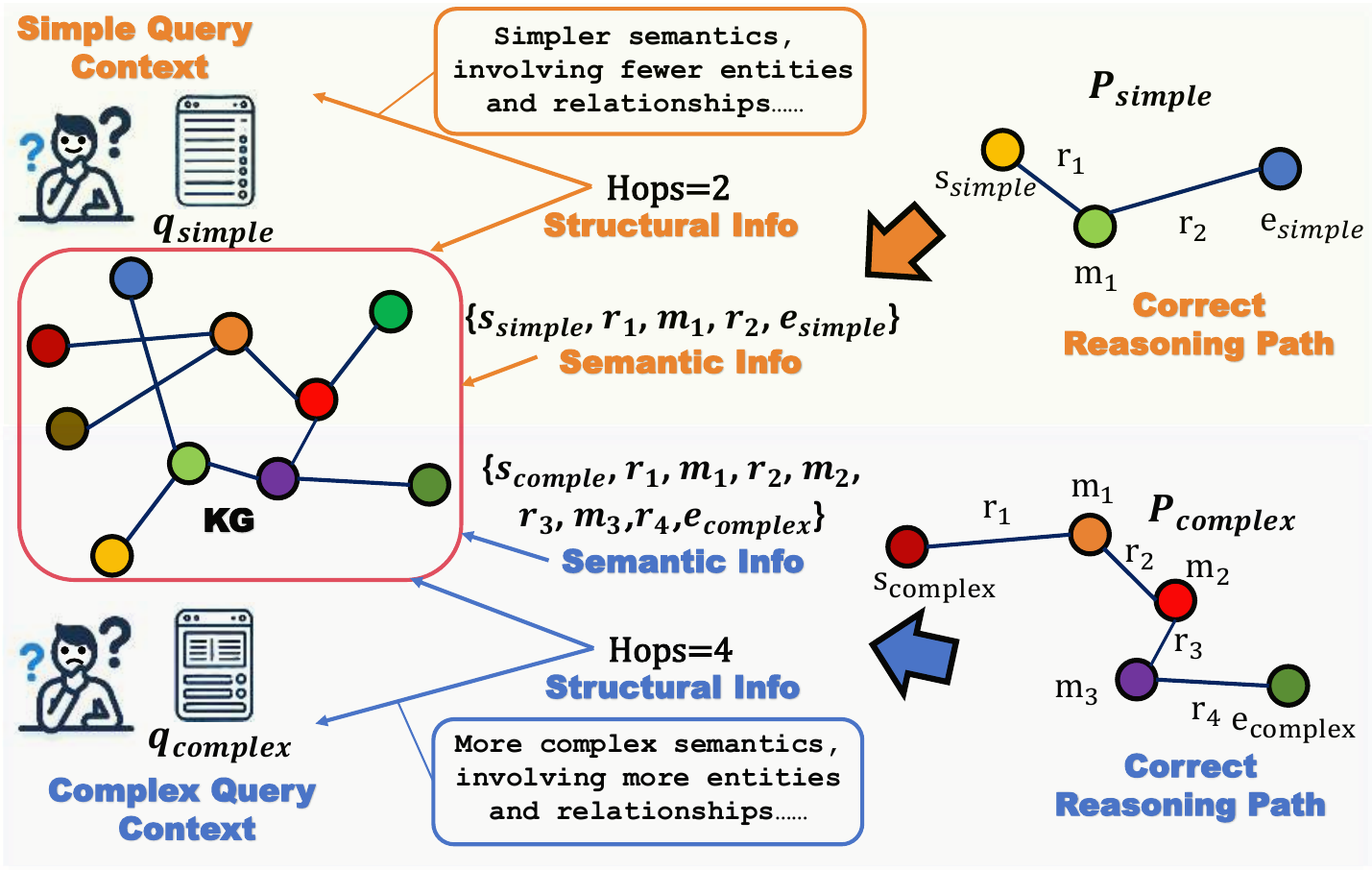}
    \caption{Analysis of Semantic and Structural Information in Reasoning Paths}
    \label{fig:twotype}
\end{figure}

In this work, we propose a novel and \underline{f}lexible modular KG-\underline{RAG} framework, called the {\it FRAG}. In a nutshell, FRAG dynamically adapts the retrieval process based on the complexity of the query context to improve reasoning accuracy, without requiring KGs information.
As shown in Figure~\ref{fig:twotype}, we analyzed the information sources associated with the correct reasoning path $P$ for a query context $q$ on the KG. A path $P$ comprises two types of information: {\it semantic} (i.e., entities and relations) and {\it structural} (i.e., number of hops). Semantic information mainly originates from KGs and is hard to perceive and utilize in advance. However, structural information is related to both KGs and the $q$ \shortcite{ChatKBQA}. Generally, the more complex the query context $q$, the greater the number of hops in the path $P$ (i.e., indicating a more difficult reasoning task) \shortcite{UHop,EmbedKGQA}. Thus, within a tolerable margin of error, we can predict the number of hops in $P$ based solely on the query context $q$. The predicted number of hops can then serve as a key factor in enhancing the non-specific retrieval process.

Starting from the above insights, the FRAG framework is primarily featured by two key modules: ``{\it Reasoning-aware}" and ``{\it Flexible-retrieval}".
The reasoning-aware module considers three key aspects of structural information prediction.
First, to reduce the impact of inherent prediction errors, it simplifies the prediction task by estimating the coarse-grained hop range, categorizing the reasoning complexity from the query context as either simple or complex based on a hop count threshold.
Second, it collects KGs and queries across various domains, extracting semantic and statistical features from the query context to train a flexible and generalizable cross-domain classifier.
Finally, an optimization strategy utilizing feedback from LLMs is employed to enhance classifier performance on specific KGs as needed.

The flexible retrieval module refines the KG-RAG retrieval process into a ``preprocessing, retrieval, and postprocessing" pipeline, facilitating the tailored customization of its three components for both simple and complex reasoning tasks.
Each component is built on traditional algorithms and models, ensuring the generality of our framework.
For simple reasoning tasks, which typically involve shorter reasoning paths, we employ breadth-first search (BFS) and ranking as the core retrieval strategies, enabling efficient and accurate retrieval. In contrast, for complex reasoning tasks characterized by longer reasoning paths, we advocate for the use of shortest path retrieval and ranking to minimize computational overhead and reduce noise, thereby improving retrieval effectiveness.


The remainder of this paper is organized as follows: In \Cref{section: related work}, we review RAG and KG-RAG frameworks. \Cref{section: preliminary} covers basic KG-RAG concepts. \Cref{section: approach} details the FRAG framework design. \Cref{section: experiment} evaluates FRAG on benchmark datasets against SOTA methods. Finally, \Cref{section: conclusion} summarizes our work and future directions.

\section{Related Work}
\label{section: related work}

{\bf Retrieval-augmented generation (RAG).}
RAG \shortcite{lewis2021retrieval} enhances LLMs by integrating retrieved knowledge during contextual learning, mitigating knowledge gaps and hallucination issues. The  NaiveRAG frameworks \shortcite{Rewriting,yu2023generateretrievelargelanguage,shao2023enhancingretrievalaugmentedlargelanguage} retrieve top-$k$ relevant documents and incorporates them into the prompt for more accurate responses. Later advancements (e.g.modular RAG \shortcite{yu2023generate, shao2023enhancing}, advanced RAG \shortcite{gao2022,peng2024large}) improved retrieval accuracy with additional modules. However, document-based RAG introduces noise and excessive context, impacting reasoning performance \shortcite{HyDE}. Recent studies \shortcite{sarmah2024} focus on altering the storage format of external knowledge. Recently, GraphRAG \shortcite{GraphRAG} unifies various knowledge  into (knowledge) graph format, transforming retrieval into a fine-grained knowledge path search, enhancing key information extraction.

    {\bf  RAG based on Knowledge Graphs (KG-RAG).}
KGs, known for dynamic, explicit, and structured knowledge representation, are increasingly used as knowledge bases for RAG. KG-RAG \shortcite{GNN-RAG, GRAG} retrieves the top-$k$ reasoning paths relevant to the query, providing concise and accurate contextual information for LLMs reasoning.
As outlined in Section 1, KG-RAG frameworks are categorized as {\it Modular} or {\it  Coupled}, based on whether KGs information is fine-tuned into the LLMs. The latest modular approach, like ToG \shortcite{ToG}, enhances retrieval accuracy by replacing traditional ranking models with LLMs. However, ToG still struggle due to a lack of prior KG knowledge and the need for frequent LLM calls. For coupled framework, the most recent method, RoG \shortcite{RoG}, fine-tunes LLMs with KG information, allowing them to generate "relation paths" as query templates that directly retrieve the correct reasoning paths from KGs. However,  RoG compromises interpretability, efficiency in knowledge updates, and the generality of the RAG process across different domain KGs.

\section{Preliminary}
\label{section: preliminary}

\textbf{Knowledge Graph.}
A Knowledge Graph (KG), $G = \{(s, r, e) \mid s, e \in V, r \in E\}$, is a structured method to represent entities ($V$) and their relationships ($E$). KGs use triples $(s, r, e)$, where $s$ and $e$ are the start and end entity, and $r$ is the relationship, to capture vast domain-specific knowledge.

\textbf{General Process of KG-RAG.}
Following existing works~\shortcite{KD-CoT,KB-BINDER,RoG,ToG,GNN-RAG}, KG-RAG involves two main stages: \textit{retrieval} and \textit{generation}. Given a query $q$, the first stage constructs a set of candidate ``reasoning paths'' by matching $q$ with entities and relationships in the KGs. This is done by searching for relevant triples $(s, r, e)$ from KG ($G$): $\textit{Retrieve}(q, G) \rightarrow \{P_i\}$. The retrieved reasoning paths set $\{P_i\}$ are then ranked by their relevance to $q$.
A reasoning path $P_i$ can be formally defined as: $P_i = (s, r_1, m_1, r_2, \ldots, r_{k-1}, m_{k-1}, r_l, e)$, where $s$ is the starting entity, $e$ is the answer entity, $m_j$s are the intermediate entities, and $r_j$s are the relationships connecting these entities. The number of hops in a path, equal to the number of relationships, determines the path's length $l$.
For a given query $q$, a correct reasoning path includes the correct answer entity corresponding to $q$. In the generation stage, the top-$k$ ranked paths augment the query, forming an enriched query $q'$. This $q'$ is then fed into LLMs to generate the final output: $\textit{Generate}(q', \text{LLM}) \rightarrow \textit{output}$.

\textbf{Simple and Complex Reasoning in KG-RAG.}
Following prior works on reasoning tasks in KGs~\shortcite{berant-etal-2013-semantic,yih-etal-2014-semantic,complex2simple,EmbedKGQA}, KG-RAG reasoning tasks can be categorized into {\it simple} and {\it complex} reasoning, based on the minimum number of hops required to find the correct answer from the KG. Given a threshold $\delta$\footnote{In this paper, we set $\delta = 2$, as most real-world reasoning tasks of KG-RAG involve paths within 2 hops \shortcite{NSM}.}, if the minimum hop count $h_{\min}$ of the reasoning path is less than or equal to $\delta$, the problem is categorized as simple; otherwise, it is classified as complex: $\textit{Type}(q) = \textit{Simple if } h_{\min}(q, G) \leq \delta \textit{ else Complex}$. For queries with multiple correct answers, the minimum hops across all correct reasoning paths are considered.
This formal classification helps in distinguishing the complexity of reasoning tasks within KG-RAG frameworks.

\section{Methodology}
\label{section: approach}
\subsection{Overview}
\begin{figure*}[t]
    \centering
    \includegraphics[width=0.99\textwidth]{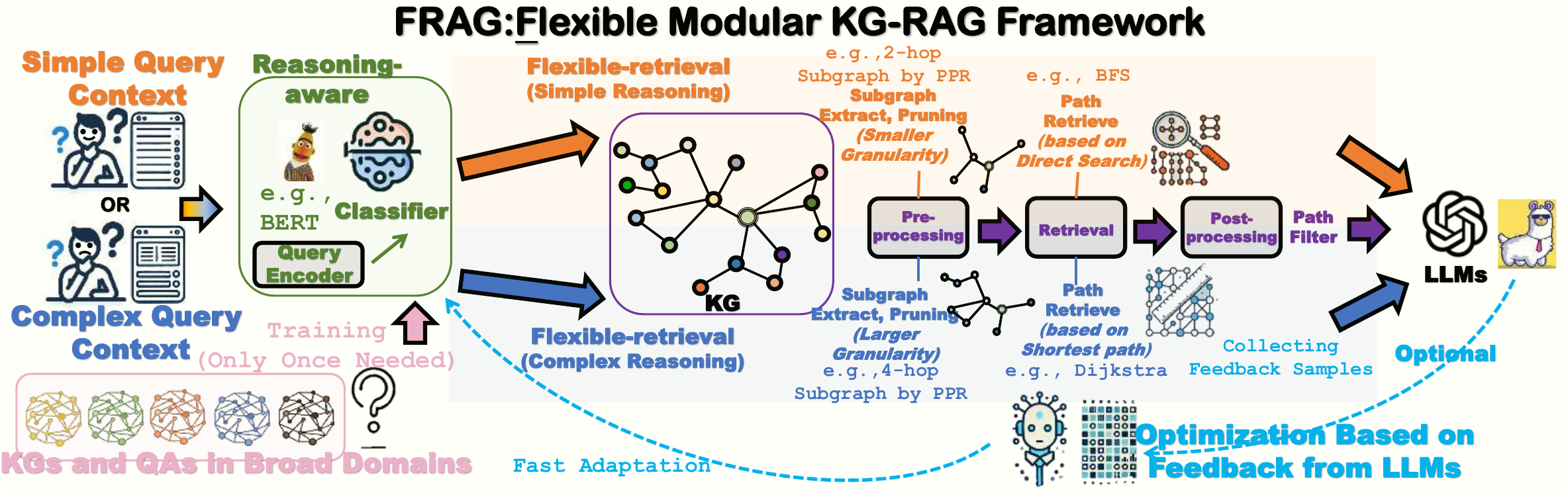}
    \caption{Framework of {\ourmethod}}
    \label{fig:c-kgrag}
\end{figure*}

As illustrated in Figure~\ref{fig:c-kgrag}, {\ourmethod} consists of three modules.
The reasoning-aware module classifies reasoning tasks as either simple or complex based on the query.
To promote the module effectiveness, we incorporate an optimization strategy that leverages the feedback from LLMs to refine the original classifier.
In the subsequent flexible-retrieval module, we refine the KG-RAG retrieval process into a ``preprocessing-retrieval-postprocessing" pipeline, tailored retrieval schemes are applies to identify accurate reasoning paths for both simple and complex tasks.
Finally, the identified reasoning paths, together with the questions, are fed to LLMs to generate answers in the reasoning module.
{\ourmethod} offers two key advantages.
First, compared to other modular approaches, {\ourmethod}'s reasoning-aware module perceives and utilizes the structural information of reasoning paths, thereby enhancing the retrieval process.
Second, by decoupling from specific KG information, the modular retrieval module grants {\ourmethod} greater generality than coupled methods.



\subsection{Reasoning-aware Module}



Identifying the complexity of a query is a prerequisite for applying targeted solutions.
For a reasoning task, {\ourmethod} classifies it as either simple or complex based on the minimum hop count of the correct reasoning paths in KGs.
Given the requirement to dissociate from the reasoning-related KG for generality, it is impractical to obtain the precise reasoning paths and, consequently, the exact number of hops.

Extensive research in knowledge graph question answering (\KGQA), along with our empirical experiments, has shown that the hop count of the correct inference path is closely linked to specific statistics in the query, such as the number of entities, relations, and clauses~\shortcite{EmbedKGQA,RnG-KBQA,Program-Transfer,RoG,ChatKBQA}.
Recognizing the relationship between the structural information of reasoning paths and the query, {\ourmethod} seeks to approximate the number of hops based solely on the query, thus categorizing the query accordingly.

To implement this, we train a binary classifier using a set of public {\KGQA} datasets, each consisting of a fundamental KG and a substantial set of queries $Q$ paired with corresponding answers $A$.
Prior to training, for each query $q\in Q$, we identify all of the shortest reasoning paths from the query entities $Ent_q$ to the answer entities in $Ent_a\in A$.
The minimum hop $\mathcal{H}$ among all reasoning paths of this query determines the query label $\mathcal{Y}$: $\mathcal{Y} = 1 \ (\text{Complex}), \text{ if } \mathcal{H} \ge \delta$; $\mathcal{Y} = 0 \ (\text{Simple}), \text{ if } \mathcal{H} < \delta$.
Here the threshold $\delta$ is set to 2 as noted earlier.
To capture the query's information, relevant statistics can be extracted and encoded.
For simplicity, the entire query is encoded as:
$
    \mathbf{h}_q = \text{QueryEncoder}(q) \in \mathbb{R}^{L_q \times d},
$
where \textit{QueryEncoder} can be any encoding mechanism, such as a language model. (e.g., BERT \shortcite{bert}), word embeddings (e.g., Word2Vec \shortcite{word2vec}), or TF-IDF.
This way, the classification loss $\mathcal{L}$ of the binary classifier during training is computed as: $\mathcal{L} = -\sum_{q\in Q}\sum_{y_q\in \mathcal{Y}}y_q\cdot \log{p(y_q|q)}$. Here, $p(y_q|q) = \text{Decoder}(\mathbf{h}_q)$ represents the probability that  $q$ is classified as either simple or complex.


Notably, unlike approaches that fine-tune LLMs or train specific models with reasoning-related KGs, which might be domain-specific or proprietary, our method leverages generic, publicly available datasets for training.
The use of reasoning-irrelevant KGs maintains the generality of our method while making the reasoning-aware module applicable to other approaches that could benefit from the idea of reasoning task classification.

Since the reasoning-aware task requires only an estimation of the hop range rather than a fine-grained perception of reasoning path, a binary classifier is sufficient for most reasoning tasks.
Moreover, we also introduce an optional approach to further optimize classification performance. This method leverages feedback from large language models (LLMs) to refine query labels, enhancing overall accuracy.
Specifically, during the reasoning phase, we prompt LLMs not only to generate standard responses, but also to identify the most relevant reasoning path, from a predefined set of input paths.
This approach allows us to derive a more accurate hop count, resulting in a refined label $ y_{fb}$ that better represents the complexity of the query.
These refined query-label pairs are collected and used to fine-tune the pre-trained binary classifier:
$
   Generate (q^\prime, \text{LLM}) \rightarrow answer, y_{fb}
$;$
    \text{Classifier} = \text{FastAdaptation} (\text{Classifier}, y_{fb})
$.
Here, $q^\prime$ is the enriched query with reasoning paths, and $y_{fb}$ is the refined label.
Such an optimization strategy requires only gathering feedbacks from the LLM, without necessitating additional fine-tuning or calls to LLMs.
The prompt used for reasoning is detailed in \textbf{Appendix}~\ref{appendix:prompt}.

\subsection{Flexible-retrieval Module}

The retrieval module aims to accurately identify reasoning paths relevant to the query from the KG.
To achieve this, we propose a {\it preprocessing-retrieval-postprocessing} pipeline.
The preprocessing step shrinks the retrieval scope, by extracting subgraphs consisting of significant entities and relations from the original KG.
During the retrieval step, tailored to the categorization of simple and complex queries, two distinct strategies are applied to search for reasoning paths.
In the postprocessing step, redundant reasoning paths are carefully filtered out to prevent the introduction of noise and unnecessary computational cost to the reasoning process.
Note that our framework is highly flexible, allowing the use of various method combinations across the three modules of the pipeline. In our implementation, we employ traditional and widely used algorithms known for their effectiveness.

\textbf{Preprocessing.}
Given a KG as $G = (V, E)$ and an entity set \( Ent_q \) from query \( q \), we extract subgraphs \( G_k^s = (V_k^s, E_k^s) \) for each entity \( s \in Ent_q \), where each subgraph \( G_k^s \) is a \( k \)-th order subgraph centered on the entity \( s \). We then take the union of these subgraphs to form the subgraph \( G_k = \bigcup_{s \in Ent_q} G_k^s = (V_k, E_k) \subseteq G \).
The parameter $k$, serving as an upper bound on the hop counts among all shortest reasoning paths, is adjusted by the complexity of the queries.

To further prune the subgraph $G_k$, we remove less relevant entities and edges based on the evaluation of their significance.
In the entity-based subgraph pruning, we employ a generalized ranking mechanism (GRM) (e.g., Random Walk with Restart (RWR); Personalized PageRank (PPR); PageRank-Nibble (PRN)), to assess the importance of the entities $v\in V_k$ relative to the query entities $Ent_q$, and then select the top $n$ entities $\widetilde{V}\subseteq V_k$:
$
    \mathbf{R}(v) = GRM(G_k, Ent_q)
$; $
    \widetilde{V} = \left\{ v_i \mid i \in \text{top-}n(\mathbf{R}(v)) \right\}
$
, where $\mathbf{R}(v)$ represents the importance score of entity $v$ relative to the query entities $Ent_q$.
The top $n$ entities $\widetilde{V}\subseteq V_k$, along with relations $\widetilde{E}\subseteq E_k$ between $\widetilde{V}$, form a subgraph $\widetilde{G}$.
Likewise, in the edge-based subgraph pruning, we apply an edge ranking model (ERM) as a retriever (e.g., BM25; SentenceTransformer), on $\widetilde{G}$ to rank the relations $r \in \widetilde{E}$ based on their semantic similarity to the query $q$:
$
    \mathbf{S}(r) = ERM(q, \widetilde{E})
$; $    \widehat{E} = \left\{ r_j \mid j \in \text{top-}m(\mathbf{S}(r)) \right\}
$
, where $\mathbf{S}(r)$ represents the similarity score of query $q$ relative to the relations set $\widetilde{E}$ of $\widetilde{G}$.
By selecting the top-$m$ relations $\widehat{E}\subseteq \widetilde{E}$ and the corresponding entities $\widehat{V}\subseteq \widetilde{V}$, we construct a more focused subgraph $\widehat{G}=(\widehat{V}, \widehat{E})$.



\textbf{Retrieval.} Upon obtaining the subgraph $\widehat{G}$ in the preprocessing, the retrieval step aims to identify reasoning paths relevant to queries on it.
Unlike {\KGQA} tasks, where reasoning paths typically contain the answer \shortcite{QGG,HGNet,Program-Transfer}, in RAG,
these paths serve as auxiliary component.
They provide value by highlighting intermediate entities and relations that may help supplement missing information in LLMs.
However, reasoning paths with excessive intermediate entities and relations introduce redundant information that can burden both retrieval and reasoning.
In other words, it introduces a trade-off between acquiring more information and maintaining efficiency.
Thus, for simple queries, which typically involve shorter reasoning paths, a broader retrieval approach is essential to minimize information loss.
Consequently, the Breadth-First Search (BFS) algorithm is employed, allowing for efficient traversal of all reasoning paths $\mathcal{P}$ between the query entities $Ent_q$ and the entities $\widehat{V}$ within $\widehat{G}$:
$
    \mathcal{P} = \{ P_i \} = \text{PathRetrieve}(\widehat{G})
$; $
    P_i =  \{ s \xrightarrow{r_1} m_1 \xrightarrow{r_2} m_2 \xrightarrow{r_3} \cdots \xrightarrow{r_k} e | s\in Ent_q, m_j\in \widehat{V}\}
$.

In contrast, for complex queries with longer reasoning paths, increasing the retrieval paths not only exponentially escalates computational cost but also introduces a large amount of redundant information. This underscores the need for efficiency and pruning.
For reasoning paths with the same start and end entities, the shortest path is preferable for directly obtaining answers~\shortcite{PullNet,EmbedKGQA,DecAF,GNN-RAG} and for reasoning with a shorter prompt.
Therefore, we resort to the Dijkstra algorithm to identify efficiently the shortest reasoning paths $\mathcal{P}$ from the query entities $Ent_q$ to entities $\widehat{V}$ within $\widehat{G}$.

\textbf{Postprocessing.} The retrieval process primarily focuses on finding paths from the query entities to potential answer entities, disregarding the semantics of intermediate entities and their relevance to the query.
This can lead to an unordered and redundant collection of reasoning paths.
Directly incorporating them into the prompt for reasoning lead to several potential issues.
First, reasoning paths that contain irrelevant or rare intermediate entities might mislead the reasoning of LLMs.
Second, these paths increase the prompt length, which not only adds to the reasoning cost but also risks exceeding the contextual length limit.
Last, the reasoning performance is influenced by the placement of these paths within the prompt, with paths positioned at the beginning having a more significant impact~\shortcite{lostInMiddle,StreamingLLM}.
To address these issues, similar to the previous method, we apply an path ranking model (PRM) (e.g., DPR \shortcite{DPR}; ColBERT \shortcite{ColBERT}; BGE \shortcite{BGE}), to rank the reasoning paths $\mathcal{P}$, based on their similarity to the query. Then, the top-$u$ reasoning paths are selected, denoted by $\mathbf{P}$:
$
    \mathbf{T}(p) = PRM(q, \mathcal{P})
$; $    \mathbf{P} = \left\{ o \mid o \in \text{top-}u(\mathbf{T}(p)) \right\}
$, where $\mathbf{T}(p)$ represents the similarity score of query $q$ relative to the reasoning paths $\mathcal{P}$.

This approach effectively filters out a substantial amount of redundant reasoning paths by leveraging the semantic correlation between the query and intermediate entities, thus shortening the prompt length and ensuring that the most relevant and beneficial reasoning paths are favorably positioned.

\subsection{Reasoning Module}

In the reasoning module, we design a prompt template to augment the question $q$ with the filtered reasoning paths $\mathbf{P}$, forming an enriched prompt $q^\prime$.
This prompt $q^\prime$ guides LLM to conduct reasoning and generate the answer:
$
    q^\prime \leftarrow prompt\, (q, \mathbf{P})
$; $
    answer   \leftarrow Generate (q^\prime, \text{LLM})
$. The reasoning prompt is detailed in \textbf{Appendix}~\ref{appendix:prompt}.

\section{Experiment}
\label{section: experiment}

\subsection{Experimental Design}

\begin{table*}[ht!]
    \centering
    \small
    \caption{Performance Comparison with Different Baselines on WebQSP and CWQ}
    \label{tab:main_exp_results}
    \resizebox{\textwidth}{!}{
    \begin{tabular}{c|p{3.5cm}@{}cc||c|p{3.5cm}@{}cc}
        \toprule
        \textbf{{Type}}                                                   & \textbf{{Methods}}                                                      & \textbf{WebQSP}                                              & \textbf{CWQ}                                            & \textbf{{Type}} & \textbf{{Methods}}                                           & \textbf{WebQSP} & \textbf{CWQ}  \\\cline{1-8}
        \multirow{14}{*}{\parbox{1.6cm}{\centering{Traditional {\KGQA}}}} & \multicolumn{3}{l||}{\textbf{\textit{Without LLMs}}}                    & \multirow{23}{*}{\parbox{1.6cm}{\centering{Modular KG-RAG}}} & \multicolumn{3}{l}{\textbf{\textit{\llamaChatFull{7}}}}                                                                                                                    \\\cline{6-8}\cline{2-4}
                                                                          & KV-Mem~\shortcite{KV-Mem}                                               & 46.7                                                         & 21.1                                                    &                 & Vanilla LLM                                                  & 63.4            & 31.1          \\
                                                                          & GraftNet~\shortcite{GraftNet}                                           & 66.4                                                         & 36.8                                                    &                 & ToG~\shortcite{ToG}                                          & 10.8            & 5.2          \\
                                                                          & PullNet~\shortcite{PullNet}                                             & 68.1                                                         & 45.9                                                    &                 & {\ourmethod} (Ours)                                          & \ul{76.6}       & \ul{47.3}     \\
                                                                          & EmbedKGQA~\shortcite{EmbedKGQA}                                         & 66.6                                                         & 45.9                                                    &                 & {\ourmethod}-F (Ours)                                        & \textbf{76.7}   & \textbf{48.9} \\\cline{6-8}
                                                                          & QGG~\shortcite{QGG}                                                     & 73.0                                                         & 44.1                                                    &                 & \multicolumn{3}{l}{\textbf{\textit{\llamaChatFull{70}}}}                                       \\\cline{6-8}
                                                                          & NSM~\shortcite{NSM}                                                     & 68.7                                                         & 47.6                                                    &                 & Vanilla LLM                                                  & 63.6            & 37.6          \\
                                                                          & TransferNet~\shortcite{TransferNet}                                     & 71.4                                                         & 48.6                                                    &                 & ToG~\shortcite{ToG}                                          & 68.9            & 57.6          \\
                                                                          & KGT5~\shortcite{KGT5}                                                   & 56.1                                                         & 36.5                                                    &                 & {\ourmethod} (Ours)                                          & \ul{81.2}       & \ul{60.1}     \\
                                                                          & SR+NSM ~\shortcite{zhang2022subgraph}                                   & 68.9                                                         & 50.2                                                    &                 & {\ourmethod}-F (Ours)                                        & \textbf{81.3}   & \textbf{62.2} \\\cline{6-8}
                                                                          & SR+NSM+E2E ~\shortcite{zhang2022subgraph}                               & 69.5                                                         & 49.3                                                    &                 & \multicolumn{3}{l}{\textbf{\textit{\llamaInstructFull{8}}}}                                    \\\cline{6-8}
                                                                          & HGNet~\shortcite{HGNet}                                                 & 70.6                                                         & \textbf{65.3}                                           &                 & Vanilla LLM                                                  & 64.0            & 37.9          \\
                                                                          & Program Transfer~\shortcite{Program-Transfer}                           & 74.6                                                         & 58.1                                                    &                 & ToG~\shortcite{ToG}                                          & 59.8            & 37.0          \\
                                                                          & UniKGQA~\shortcite{UniKGQA}                                             & \textbf{77.2}                                                & 51.2                                                    &                 & {\ourmethod} (Ours)                                          & \ul{87.7}       & \ul{64.9}     \\\cline{1-4}
        \multirow{3}{*}{\parbox{1.6cm}{\centering{Coupled KG-RAG}}}       & \multicolumn{3}{l||}{\textbf{\textit{{\llamaChatFull{7}} (Finetuned)}}} &                                                              & {\ourmethod}-F (Ours)                                   & \textbf{87.8}   & \textbf{66.1}                                                                                  \\\cline{2-4}\cline{6-8}
                                                                          & GNN-RAG~\shortcite{GNN-RAG}                                             & 80.6                                                         & 61.7                                                    &                 & \multicolumn{3}{l}{\textbf{\textit{\llamaInstructFull{70}}}}                                   \\\cline{6-8}
                                                                          & RoG~\shortcite{RoG}                                                     & \textbf{85.7}                                                & \textbf{62.6}                                           &                 & Vanilla LLM                                                  & 73.1            & 46.1          \\\cline{1-4}
        \multirow{6}{*}{\parbox{1.6cm}{\centering{Modular KG-RAG}}}       & \multicolumn{3}{l||}{\textbf{\textit{ChatGPT}}}                         &                                                              & {\ourmethod} (Ours)                                     & \textbf{88.6}   & \ul{69.4}                                                                                      \\\cline{2-4}
                                                                          & Vanilla LLM                                                             & 66.8                                                         & 39.9                                                    &                 & {\ourmethod}-F (Ours)                                        & \textbf{88.6}   & \textbf{70.8} \\\cline{6-8}
                                                                          & KD-CoT~\shortcite{KD-CoT}                                               & 73.7                                                         & 50.5                                                    &                 & \multicolumn{3}{l}{\textbf{\textit{GPT-4o-mini}}}                                              \\\cline{6-8}
                                                                          & ToG~\shortcite{ToG}                                                     & 76.2                                                         & 57.1                                                    &                 & Vanilla LLM                                                  & 69.2            & 43.8          \\
                                                                          & {\ourmethod} (Ours)                                                     & \textbf{82.3}                                                & \ul{61.0}                                               &                 & {\ourmethod} (Ours)                                          & \textbf{86.7}   & \ul{66.9}     \\
                                                                          & {\ourmethod}-F (Ours)                                                   & \textbf{82.3}                                                & \textbf{61.5}                                           &                 & {\ourmethod}-F (Ours)                                        & \textbf{86.7}   & \textbf{68.0} \\
        \bottomrule
    \end{tabular}
    }
\end{table*}

\textbf{Datasets and Evaluation Metrics.}
Our experiments utilize two widely recognized {\KGQA} datasets: WebQSP~\shortcite{webqsp} and CWQ~\shortcite{CWQ}, both extensively used in the KGQA and KG-RAG research community~\shortcite{KD-CoT,DecAF,RoG,GNN-RAG,ChatKBQA,ToG}. 
Table~\ref{tab:hop_dist} summarizes the distribution of question hops across these datasets, showing that both predominantly feature simple queries. Specifically, WebQSP consists entirely of simple queries, while CWQ includes a small fraction (20.75\%) of complex ones. Further dataset details are provided in \textbf{Appendix}~\ref{appendix:datasets}.
Following prior works~\shortcite{ knowledgeaugmented, structgpt, CoK,KB-BINDER, ToG, ToG2.0}, we employ Hits@1 score as the evaluation metric, assessing the percentage of correct answers ranked first by LLM.


\begin{table}[t]
    \centering
    \small
    \caption{Statistics of Question Hops of Datasets}
    \label{tab:hop_dist}
    \begin{tabular}{@{}cccc@{}}
        \toprule
        Dataset & 1 hop                         & 2 hop   & $\geq$ 3 hop \\ \midrule
        WebQSP  & 65.49                      \% & 34.51\% & 0.00\%       \\
        CWQ     & 40.91                      \% & 38.34\% & 20.75\%      \\ \bottomrule
    \end{tabular}%
\end{table}


\textbf{Reasoning LLMs.}
We evaluate the performance of our approach using six LLMs, which are referenced in this paper as follows:
Llama-2-7b-chat-hf (\llamaChat{7}), Llama-2-70b-chat-hf (\llamaChat{70}) \shortcite{llama2}, Llama-3-8B-Instruct (\llamaInstruct{8}), Llama-3-70B-Instruct (\llamaInstruct{70}) \shortcite{llama3}, GPT-3.5-turbo (ChatGPT) and GPT-4o-mini.

\textbf{Baselines.}
Given that {\ourmethod} is a modular KG-RAG approach, we primarily compare it with other modular KG-RAG methods, as well as with coupled KG-RAG and traditional {\KGQA} techniques.
Among them,
{\bf \textit{1) Modular KG-RAG:}} Vanilla LLMs (without RAG), KD-CoT \shortcite{KD-CoT} and ToG \shortcite{ToG}. KD-CoT enhances CoT prompting by integrating KG knowledge. ToG leverages LLMs to iteratively select the most pertinent relations and entities, representing the current SOTA in modular KG-RAG methods. 
    {\bf \textit{2) Coupled KG-RAG:}} GNN-RAG \shortcite{GNN-RAG} and RoG \shortcite{RoG}. GNN-RAG leverages GNN to extract useful reasoning paths. RoG utilizes a fine-tuned LLM to generate relation paths for answering questions, representing the SOTA in coupled KG-RAG. 
{\bf \textit{3) Traditional {\KGQA}:}} 13 traditional {\KGQA} methods, as described in \textbf{Appendix}~\ref{appendix:baseline}.

\textbf{Experiment Implementations.}
{\bf \textit{1) For the reasoning-aware module}}, we employ DeBERTaV3~\shortcite{debertav3} as the query encoder and decoder.
We construct training datasets for simple and complex reasoning tasks using two large and cross-domain KG databases, Freebase~\shortcite{Freebase} and Wiki-Movies~\shortcite{MetaQA}, and ensuring complete isolation between the training data and the two test datasets.
    {\bf \textit{ 2) For the flexible-retrieval  module}}, we utilize bge-reranker-v2-m3~\shortcite{BGE} as the \(ERM\) and \(PRM\) during both preprocessing and postprocessing stages. In the preprocessing stage, we use PPR algorithm as \(GRM\), and the hyperparameters as follows: \( k=2 \), \( n=2000 \), \( m=64 \) for simple queries, and \( k=4 \), \( n=2000 \), \( m=64 \) for complex queries. In the postprocessing stage, we set \( u=32 \) across all experiments.
    {\bf \textit{ 3) For LLMs' reasoning}}, we use zero-shot prompting to LLMs generation.
    {\bf \textit{ 4) Feedback adjustment is optional}}, with the adjustment rate is set to 0.25, indicating that {25\%} of the samples are selected for fast adaptation of the reasoning-aware module.
Detailed settings are provided in \textbf{Appendix}~\ref{appendix:exp_setting}.

\subsection{Results}

\begin{figure}[ht]
    \centering
    \includegraphics[width=0.49\textwidth]{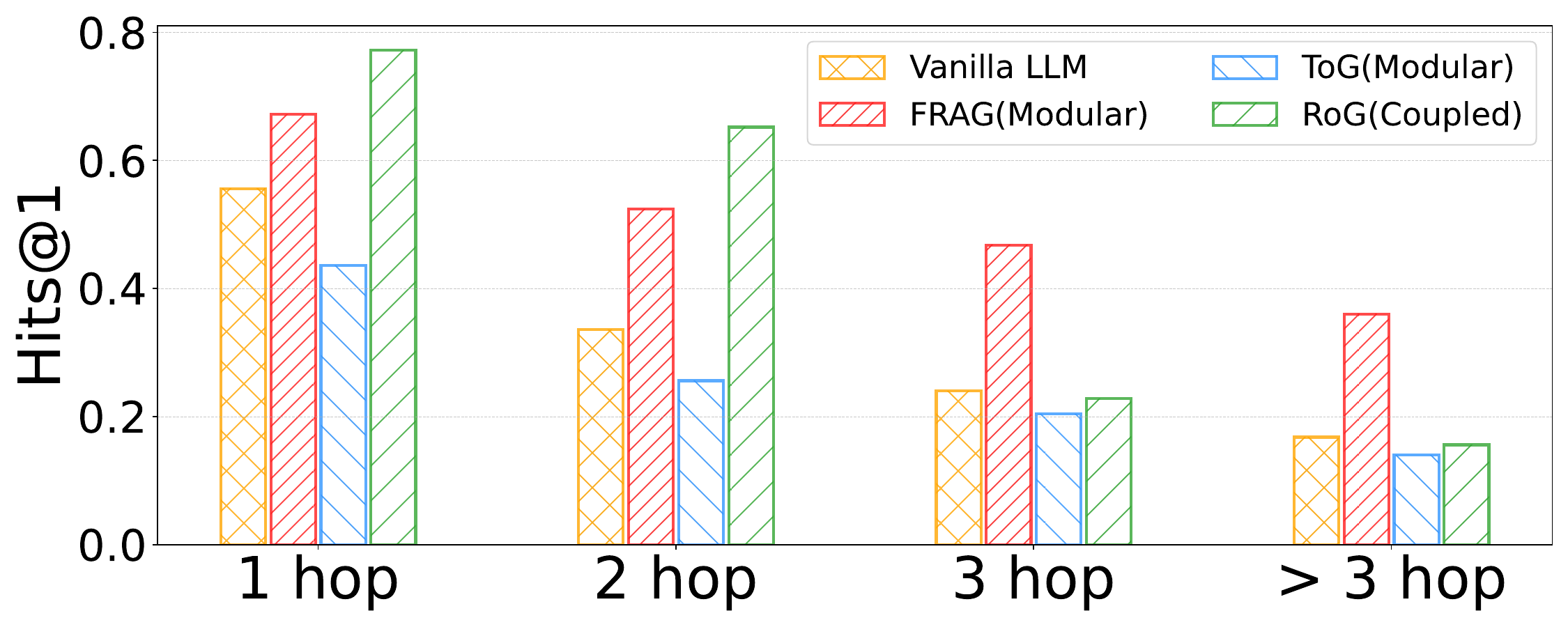}
    \caption{Performance of Different Hops (\llamaChat{7})}
    \label{fig:hop_llama2}
\end{figure}

Table~\ref{tab:main_exp_results} presents a detailed performance comparison of our method against various baselines, while Figure~\ref{fig:hop_llama2} further illustrate detailed evaluations of sampled queries across different hops, both demonstrating the effectiveness of \ourmethod. Additionally, Table~\ref{tab:training-time} compare the cost of {\ourmethod} with ToG and RoG in terms of training and retrieval operations. The following analysis will highlight our method's advantages in both effectiveness and efficiency.

\textbf{FRAG Achieves New SOTA as a Modular KG-RAG.}
As shown in Table~\ref{tab:main_exp_results}, our method consistently outperforms the previous SOTA, ToG, across various settings, establishing itself as the most advanced modular KG-RAG method. For example, on the Llama-3-8B-Instruct model, FRAG improves the scores from 59.8 and 37.0 to 87.7 and 64.9, respectively, achieving a 46.7\% and 75.4\% improvement on two datasets compared to ToG. Furthermore, incorporating feedback in FRAG-F further elevates its performance to 81.3 and 62.2.
Notably, ToG exhibits performance degradation on small scales (7B and 8B) LLM bases, such as Llama-2-7b-chat-hf, where the scores on the two datasets drop from 63.4 and 31.1 to 10.8 and 5.2, respectively, compared to the vanilla LLM. In contrast, our method reliably enhances output quality through KG RAG across all LLMs. Additionally, Figure~\ref{fig:hop_llama2} further demonstrates the robust performance of FRAG, which delivers significant improvements over both ToG and vanilla LLMs in varied hop numbers.

Moreover, ToG iteratively select the most relevant relations and entities, resulting in a significantly higher average number of LLM calls compared to FRAG's zero LLM call. This underscores the efficiency of our approach, which delivers superior performance with minimal  overhead.

\textbf{FRAG Achieves Comparable Performance with Lower Training Costs.}
Unlike traditional {\KGQA} methods that rely on specific models to embed KG semantic information, FRAG leverages pretrained LLMs with generalizable retrieval and filtering modules, achieving superior performance with minimal effort. Even with the small-parameter Llama-3-8B-Instruct, our method outperforms all traditional {\KGQA} methods. Compared to the coupled KG-RAG approach, like advanced RoG, FRAG delivers similar performance while drastically reducing training and fine-tuning time. As shown in Table~\ref{tab:training-time}, RoG requires 38 hours to fine-tune a 7B LLM, whereas FRAG only considers the structure knowledge of KG in just 306 seconds of training, but reaching approximately 89\% of RoG's performance on WebQSP. Moreover, FRAG’s plug-and-play nature allows seamless enhancement with larger-scale LLMs like Llama-3-70B-Instruct, achieving scores of 88.6 and 69.4 on two datasets. This kind of enhancement would be challenging for RoG due to the significantly higher fine-tuning costs. Additionally, Figure~\ref{fig:hop_llama2} highlights FRAG’s superior performance in complex reasoning scenarios involving more than two hops, outperforming RoG in these tasks.

\begin{table}[t]
    \small
    \caption{Training and Retrieval Cost Comparison}
    \label{tab:training-time}
    \resizebox{\columnwidth}{!}{
    \begin{tabular}{@{}c|c@{}cc|c@{}}
        \toprule
        \multirow{3}{*}{Method} & \multicolumn{3}{c|}{Training Cost (Time)}                               & \multirow{3}{*}{\begin{tabular}[c]{@{}c@{}}Retrieval Cost\\ (Ave. LLM Calls)\end{tabular}}              \\ \cline{2-4}
                                & \multicolumn{2}{c|}{Reasoning-aware}                                    & \multirow{2}{*}{Fine-tune}                                                                 &            \\ \cline{2-3}
                                & Training                                                                & \multicolumn{1}{c|}{Feedback}                                                              &     &      \\ \midrule
        ToG                     & -                                                                       & -                                                                                          & -   & 13.3 \\
        RoG                     & -                                                                       & -                                                                                          & 38h & 3    \\
        \midrule
        FRAG                    & 306s{\tablefootnote{Only once for the Reasoning-aware module training}} & -                                                                                          & -   & 0    \\
        FRAG-F                  & -                                                                       & 7.58s                                                                                      & -   & 0    \\
        \bottomrule
    \end{tabular}%
    }
\end{table}

\subsection{Ablation Study}

\begin{table}[t]
    \small
    \caption{Ablation Study Results on Two Datasets}
    \label{tab:ablation_results}
    \resizebox{\columnwidth}{!}{
    \begin{tabular}{@{}c|c@{\hskip 3pt}c|c@{\hskip 3pt}c|c@{\hskip 2pt}c@{}}
        \toprule
        \multirow{2}{*}{Method} & \multicolumn{2}{c|}{Llama2} & \multicolumn{2}{c|}{Llama3} & \multicolumn{2}{c}{GPT}                                                 \\\cmidrule{2-7}
                                & 7B                          & 70B                         & 8B                      & 70B           & 3.5-turbo     & 4o-mini       \\\midrule
                                & \multicolumn{6}{c}{CWQ}                                                                                                             \\\cmidrule{2-7}
        FRAG-F                  & \textbf{48.9}               & \textbf{62.2}               & \textbf{66.1}           & \textbf{70.8} & \textbf{61.5} & \textbf{68.0} \\
        FRAG                    & \ul{47.3}                   & 60.1                        & \ul{64.9}               & \ul{69.4}     & \ul{61.0}     & 66.9          \\
        FRAG-Simple             & 47.1                        & \ul{60.8}                   & 63.2                    & 68.6          & 59.5          & \ul{67.4}     \\
        FRAG-Complex            & 47.0                        & 59.5                        & 62.5                    & 66.6          & 59.7          & 64.9          \\
        Vanilla LLM             & 31.1                        & 37.6                        & 37.9                    & 46.1          & 39.9          & 43.8          \\ \midrule
                                & \multicolumn{6}{c}{WebQSP}                                                                                                          \\\cmidrule{2-7}
        FRAG-F                  & \ul{76.7}                   & \textbf{81.3}               & \textbf{87.8}           & \textbf{88.6} & \textbf{82.3} & \textbf{86.7} \\
        FRAG                    & 76.6                        & 81.2                        & \ul{87.7}               & \textbf{88.6} & \textbf{82.3} & \textbf{86.7} \\
        FRAG-Simple             & \textbf{76.8}               & \textbf{81.3}               & \ul{87.7}               & \textbf{88.6} & \textbf{82.3} & \textbf{86.7} \\
        FRAG-Complex            & 72.0                        & 76.4                        & 80.4                    & 81.7          & 76.2          & 81.5          \\
        Vanilla LLM             & 63.4                        & 63.6                        & 64.0                    & 73.1          & 66.8          & 69.2          \\\bottomrule
    \end{tabular}%
    }
\end{table}

We conduct ablation experiments to compare our method with those using only the simple reasoning pipeline ({\ourmethod}-Simple) or only the complex reasoning pipeline ({\ourmethod}-Complex). As shown in Table~\ref{tab:ablation_results}, without the reasoning-aware module that routes simple and complex queries through distinct pipelines, performance slightly declines when a single reasoning approach is applied to all questions. This is partly due to the limited presence of complex queries in the CWQ dataset (20.75\%) and their absence in the WebQSP dataset.
Notably, although the WebQSP dataset does not contain complex queries, {\ourmethod} did not lead to a significant performance decline compared to {\ourmethod}-Simple. This further substantiates that our method can intelligently allocate the appropriate pipeline for each type of question, ensuring optimal performance.
To further explore the impact of our proposed modules under more balanced conditions, we conducts additional ablation experiments on datasets with a more equal distribution of simple and complex queries. These results are provided in the \textbf{Appendix}~\ref{appendix:ablation}.

\section{Conclusion}
\label{section: conclusion}

In this paper, we propose {\ourmethod}, a modular KG-RAG framework that addresses the challenge of enhancing reasoning accuracy in LLMs without compromising flexibility. By adapting the retrieval process based on the complexity of the query context, FRAG leverages structural information predictions to optimize retrieval strategies. Thus, FRAG comprises two key modules: the reasoning-aware module predicts the complexity of the reasoning tasks (simple or complex) based solely on the query context, while the flexible-retrieval module customizes the retrieval process according to task complexity to enhance retrieval efficiency and effectiveness. Extensive experiments show that FRAG improves retrieval quality while maintaining flexibility, outperforming  existing KG-RAG approaches. In the future, we aim to further enhance {\ourmethod}'s adaptability to more diverse knowledge graph structures and complex reasoning scenarios.


\section*{Limitations}

FRAG represents a highly flexible framework, yet its potential and adaptability remain underexplored in the current experiments, which primarily rely on widely adopted algorithms and models. For instance, the reasoning-aware module employs DeBERTaV3, the generalized ranking mechanism utilizes PPR, and both the edge and path ranking models adopt bge-reranker-v2-m3. In diverse scenarios, these modules can be substituted with alternative implementations, such as BM25 or large language models (LLMs), to achieve more effective trade-offs between performance and computational efficiency. This adaptability highlights FRAG's capacity to be tailored to varying requirements across different applications.


\bibliography{FRAG.bib}

\clearpage
\appendix

\newpage

\section{Appendix}

\subsection{Algorithm for FRAG}
\label{appendix:algorithm}

\begin{algorithm}[ht!]
    \KwData{Knowledge graph $G = (V, E)$, Dataset $\mathcal{D}$, Feedback flag $flag$, adjustment rate $ratio$;}
    \KwResult{Reasoning result $A$;}

    $N \gets $ size of $\mathcal{D}$\;
    $i \gets 0$\; 
    Initialize list of answers $\mathcal{A} \gets [\ ]$\;
    $E$=extractEntities($q$)\;
    \ForEach {query $q$ in Dataset $\mathcal{D}$}{
        Classifier($q$)\;
        $G_k$ $ \leftarrow $ SubgraphExtract($G, q$)\;
        \If{$q$ is simple query}{
            $\widetilde{G}$ $ \leftarrow  GRM_{s}$ ($G_k, E, n_{s}, k_{s}$)\;
            $\widehat{G}$ $ \leftarrow  ERM_{s}$ ($\widetilde{G}, q, m_{s}$)\;
            $\mathcal{P}$ $ \leftarrow PathRetrieve_{s}$ ($\widehat{G}$)\;
            $P$ $ \leftarrow  PRM_{s}$ ($\mathcal{P}, u_{s}$)\;
        }
        \ElseIf{$q$ is complex query}{
            $\widetilde{G}$ $ \leftarrow GRM_{c}$ ($G, E, n_{c}, k_{c}$)\;
            $\widehat{G}$ $ \leftarrow ERM_{c}$ ($\widetilde{G}, q, m_{c}$)\;
            $\mathcal{P}$ $ \leftarrow PathRetrieve_{c}$ ($\widehat{G}$)\;
            $P$ $ \leftarrow PRM_{c}$($\mathcal{P}, u_{c}$)\;
        }

        $ans, y_{fb}$ $ \leftarrow $ Generate(LLM, concat($P, q$))\;

        \If{$flag$ is True \textbf{and} $i<ratio \cdot N$}{
            FastAdaptation(Classifier, $q$, $y_{fb}$)\;
        }

        $i \gets i + 1$\; 
        Add $ans$ to $\mathcal{A}$\;
    }
    \Return{$A$}\;
    \caption{{\ourmethod} Framework}\label{workflow}
\end{algorithm}

\begin{algorithm}[ht!]
    \SetAlgoLined
    \KwData{Knowledge Graph $G$, Start node $s$}
    \KwResult{All paths $\mathcal{P}$ from $s$ to every other node in $G$}

    Initialize queue $Q \gets \{(s, [s])\}$\;
    Initialize list of paths $\mathcal{P} \gets [\ ]$\;

    \While{$Q$ is not empty}{
    Dequeue the first element $(v, path)$ from $Q$\;
    \ForEach{neighbor $u$ of $v$}{
    \If{$u \notin path$}{
    Enqueue $(u, path + [u])$ to $Q$\;
    Add $path + [u]$ to $\mathcal{P}$\;
    }
    }
    }

    \Return $\mathcal{P}$\;
    \caption{Retrieval Algorithm (FRAG-Simple)}\label{bfs}
\end{algorithm}

\begin{algorithm}[ht!]
	\SetAlgoLined
	\KwData{Knowledge Graph $G = (V, E)$, Start node $s\ (s \in V)$}
	\KwResult{Set of shortest paths $\mathcal{P}$ from $s$ to all other nodes in $G$}
	
	Initialize distances: $d[v] \gets \infty$ for all $v \in V$, $d[s] \gets 0$\;
	Initialize priority queue $Q$ as a min-heap\;
	Insert $s$ into $Q$ with priority $d[s]$\;
	Initialize predecessor array $pred[v] \gets \text{null}$ for all $v \in V$\;
	
	\While{$Q$ is not empty}{
		Extract node $u$ from the top of $Q$\;
		\ForEach{neighbor $v$ of $u$}{
			\If{$d[u] + 1 < d[v]$}{
				$d[v] \gets d[u] + 1$\;
				$pred[v] \gets u$\;
				\If{$v$ is not in $Q$}{
					Insert $v$ into $Q$ with priority $d[v]$\;
				}
				\Else{
					Update priority of $v$ in $Q$ to $d[v]$\;
				}
			}
		}
	}
	
	Initialize shortest paths set $\mathcal{P} \gets \{\}$\;
	\ForEach{node $t \in V \setminus \{s\}$}{
		Initialize path list $path \gets [\ ]$\;
		$u \gets t$\;
		\While{$u \neq \text{null}$}{
			Prepend $u$ to $path$\;
			$u \gets pred[u]$\;
		}
		Add $path$ to $\mathcal{P}$\;
	}
	\Return $\mathcal{P}$\;
	\caption{Retrieval Algorithm (FRAG-Complex)}\label{dijkstra}
\end{algorithm}


\subsection{Datasets}
\label{appendix:datasets}
\begin{table}[!btp]
    \centering
    \caption{Statistics of datasets.}
    \label{tab:datasets}
    \begin{tabular}{@{}c|ccc@{}}
        \toprule
        Datasets & Train  & Test  & Max hop \\
        \midrule
        WebQSP   & 2,826  & 1,628 & 2       \\
        CWQ      & 27,639 & 3,531 & 4       \\
        \bottomrule
    \end{tabular}
\end{table}

\begin{table}[hbp]
    \centering
    \caption{Statistics of the Number of Answers}
    \label{tab:ans_dist}
    \begin{tabular}{@{}ccccc@{}}
        \toprule
        Dataset & \#Ans = 1 & 2--4   & 5--9  & $\geq$ 10 \\ \midrule
        WebQSP  & 51.2\%    & 27.4\% & 8.3\% & 12.1\%    \\
        CWQ     & 70.6\%    & 19.4\% & 6\%   & 4\%       \\ \bottomrule
    \end{tabular}
\end{table}

\begin{table}[!btp]
    \centering
    \caption{Statistics of Sample Size of KG-1000}
    \label{tab:sample_dataset}
    \begin{tabular}{@{}c|cccc@{}}
        \toprule
        Datasets & 1-hop & 2-hop & 3-hop & 4-hop \\
        \midrule
        KG-1000  & 250   & 250   & 250   & 250   \\
        \bottomrule
    \end{tabular}
\end{table}
We adopt two benchmark {\KGQA} datasets: WebQuestionSP (WebQSP)~\cite{webqsp} and Complex WebQuestions (CWQ)~\cite{CWQ}. We follow previous works~\cite{ToG, RoG, GNN-RAG} to use the same train and test splits for fair comparison. The statistics of the datasets are shown in Table~\ref{tab:datasets}. The distribution of the answer numbers are shown in Table~\ref{tab:ans_dist}.

To ensure a rigorous and balanced evaluation of our proposed method, we constructs additional datasets with an equal distribution of simple and complex queries. Specifically, we randomly samples 1,000 instances from the CWQ and WebQSP datasets to form the KG-1000 subsets, where the proportion of queries for each hop was meticulously balanced. This carefully designed sampling strategy allows us to evaluate the effectiveness of the reasoning-aware module under more balanced conditions, thus providing a more precise assessment of its impact on performance across varying query types.

\subsection{Baselines}
\label{appendix:baseline}
Below, we introduce 13 traditional {\KGQA} methods in order of publication, as they were not covered in detail earlier.
\begin{itemize}
    \item KV-Mem~\cite{KV-Mem} is a key-value structured memory network to retrieve answers from KGs.
    \item GraftNet~\cite{GraftNet} is a graph convolution-based neural network that reasons over KGs.
    \item PullNet~\cite{PullNet} extends GraftNet by iteratively constructing a question-specific subgraph to facilitate reasoning.
    \item EmbedKGQA~\cite{EmbedKGQA} models the reasoning on KGs through the embeddings of entities and relations.
    \item QGG~\cite{QGG} proposes a segmented query graph generation method that flexibly generates query graphs by simultaneously incorporating constraints and extending relationship paths.
    \item NSM~\cite{NSM} introduces a teacher-student framework to simulate the multi-hop reasoning process.
    \item TransferNet~\cite{TransferNet} implements a graph neural network to effectively capture the relationship between entities and questions, enabling reasoning within a unified framework that handles both label and text relations.
    \item KGT5~\cite{KGT5} leverages a fine-tuned sequence-to-sequence model on knowledge graphs to generate answers directly from the input question.
    \item SR+NSM\cite{zhang2022subgraph} introduces a method for multi-hop reasoning that retrieves relevant subgraphs through a relation-path retrieval mechanism.
    \item SR+NSM+E2E\cite{zhang2022subgraph} enhances SR+NSM by employing an end-to-end approach that jointly optimizes both the retrieval and reasoning components.
    \item HGNet~\cite{HGNet} introduces a hierarchical approach for generating query graphs, which includes an initial outlining stage to establish structural constraints, followed by a filling stage focused on selecting appropriate instances.
    \item Program Transfer~\cite{Program-Transfer} presents a two-stage parsing framework for complex \KGQA, utilizing an ontology-guided pruning strategy.
    \item UniKGQA~\cite{UniKGQA} unifies retrieval and reasoning using a single retriever-reader model.
\end{itemize}


\subsection{Experiment Detail}
\label{appendix:exp_detail}
\subsubsection{Detailed Experimental Settings.}
\label{appendix:exp_setting}
All experiments are running on Ubuntu 20.04.6 LTS (Intel(R) Xeon(R) Platinum 8358 CPU@2.60GHz Processor, 4 A100-80G, 400GB memory). The detailed experimental settings are as follows:
{\bf \textit{1) Reasoning-aware module.}}
The reasoning-aware module underwent training on the designated datasets for 10 epochs with a batch size of 32. The learning rate is set to 1e-5, and the weight decay parameter is set to 0.
{\bf \textit{2) Retrieval.}}
The damping factor \( \alpha \) of PPR algorithm is set to 0.8, and the maximum iteration is set to 1000.
{\bf \textit{3) Reasoning.}}
The temperature parameter is set to 0.01, and the maximum token length for generation is fixed at 256. We use zero-shot reasoning prompt across all datasets, and the prompt templates are presented in \textbf{Appendix}~\ref{appendix:prompt}.
{\bf \textit{4) Feedback Adjustment.}}
After collecting the feedback datasets, the reasoning-aware module is trained for 3 epochs with the batch size of 16. The learning rate is fixed at 1e-4, and the weight decay parameter is set to 0.

\subsubsection{Ablation Study on KG-1000.}
\label{appendix:ablation}
\begin{table}[h!]
    \centering
    \small
    \caption{Ablation Study Results on KG-1000}
    \label{tab:ablation_results_KG-1000}
    \begin{tabular}{@{}c|c|c@{}}
        \toprule
        {Method}    & {\llamaChat{7}} & {\llamaInstruct{8}} \\
        \midrule
        \ourmethod         & \textbf{50.6}              & \textbf{63.2}                  \\
        \ourmethod-Simple  & 40.3                       & 49.6                           \\
        \ourmethod-Complex & 46.9                       & 59.8                           \\
        Vanilla LLM        & 32.6                       & 38.4                           \\
        \bottomrule
    \end{tabular}
\end{table}
We conduct an ablation study on the KG-1000 dataset to evaluate the effectiveness of the reasoning-aware module with a more equal distribution of simple and complex queries. As shown in Table~\ref{tab:ablation_results_KG-1000}, without the reasoning-aware module, the performance of the model drops significantly.

\subsubsection{Wilcoxon signed-rank tests.}
\label{appendix:exp_result}
\begin{table}[!hbtp]
    \centering
    \caption{P-Value and Statistic (Wilcoxon Signed-rank Test)}
    \label{tab:Wilcoxon}
    \begin{tabular}{@{}c|cc@{}}
        \toprule
        Method                & p-value & statistic \\
        \midrule
        FRAG v.s. ToG         & 1.95E-3 & 0         \\
        FRAG v.s. vanilla LLM & 4.88E-4 & 0         \\
        \bottomrule
    \end{tabular}
\end{table}
We conduct Wilcoxon signed-rank tests to evaluate the statistical significance of the improvements achieved by {\ourmethod} over the baselines. As shown in Table~\ref{tab:Wilcoxon}, all p-values are below 0.05 threshold, indicating that the improvements achieved with {\ourmethod} are statistically significant.

\subsection{Prompts}
\label{appendix:prompt}
The zero-shot reasoning prompt for reasoning-aware module is as follow:
\begin{tcolorbox}[colframe=black, colback=white, boxrule=0.4mm, rounded corners, arc=0mm, left=2mm, right=2mm, top=2mm, bottom=2mm]
    \noindent\textbf{Prompt:} You are an expert reasoner with a deep understanding of logical connections and relationships. Your task is to analyze the given reasoning paths and provide accurate reasoning path to the questions based on these paths. Based on the reasoning paths, please extract the correct reasoning path. If NO correct reasoning path, please just reply NO.

    Reasoning Paths: \{paths\} \quad

    Question: \{question\} \quad

    Correct reasoning path:
\end{tcolorbox}

The zero-shot reasoning prompt for reasoning module is as follow:
\begin{tcolorbox}[colframe=black, colback=white, boxrule=0.4mm, rounded corners, arc=0mm, left=2mm, right=2mm, top=2mm, bottom=2mm]
    \noindent\textbf{Prompt:} You are an expert reasoner with a deep understanding of logical connections and relationships. Your task is to analyze the given reasoning paths and provide clear and accurate answers to the questions based on these paths. Based on the reasoning paths, please answer the given question.

    Reasoning Paths: \{paths\}

    Question: \{question\}
\end{tcolorbox}

\end{document}